\documentclass[11pt]{llncs}
    
    \usepackage{llncsdoc}

    \usepackage{comment,algorithmic,algorithm,float}
    \usepackage{amsfonts} 
    \usepackage{amssymb}
    \usepackage{hhline}
    \usepackage{enumitem}
    \usepackage{authblk}
    \usepackage{amsmath} 
    \usepackage{color}
    \usepackage{stmaryrd}
    \usepackage{soul}
    \usepackage{times}
    \usepackage{epsfig}
    
    \DeclareMathOperator*{\argmax}{argmax}
    \usepackage{subcaption}
    \usepackage{array}
    \newcolumntype{C}[1]{>{\centering\arraybackslash}m{#1}}
    \usepackage{comment,float}
    \usepackage[latin1]{inputenc}
    \usepackage{enumitem}
    \usepackage{authblk}
    \usepackage{multirow}
    \usepackage{fontenc}
    \usepackage{calc}
    \usepackage{indentfirst}
    \usepackage{fancyhdr}
    \usepackage{graphicx}
    \usepackage{epstopdf}
    \usepackage{lastpage}
    \usepackage{ifthen}
    \usepackage{lineno}
    \usepackage{float}
    \usepackage{setspace}
    \usepackage{mathpazo}
    \usepackage{booktabs}
    \let\llncssubparagraph\subparagraph
    \let\subparagraph\paragraph
    \usepackage[compact]{titlesec}
    \let\subparagraph\llncssubparagraph
    \usepackage{etoolbox}
    \usepackage{tabto}
    \usepackage{xcolor}
    \usepackage{soul}
    \usepackage{microtype}
    \usepackage{tikz}
    \usepackage{totcount}
    \usepackage{amsthm}
    \usepackage{hyphenat}
    \usepackage{footmisc}
    \usepackage{url}
    \usepackage{geometry}
    \usepackage{newfloat}
    \usepackage{caption}
    \usepackage[toc,page]{appendix}

    \usepackage{times}
    \usepackage{epsfig}
    \usepackage{graphicx}
    \usepackage{color}

    \usepackage{subcaption}
    
    \title{Probabilistic Jacobian-based Saliency Maps Attacks}
    
    \author{Th\'eo Combey $^{1}$, Ant\'onio Loison $^{1}$, Maxime Faucher $^{1}$ and Hatem Hajri $^{2}$}

    \institute{CentraleSup\'elec, 3 Rue Joliot-Curie 91192, Gif-sur-Yvette, France,\\
    \email{theo.combey, maxime.faucher@student-cs.fr}
    \and
    IRT SystemX, 8 Avenue de la Vauve, 91120 Palaiseau, France\\
    \email{hatem.hajri@irt-systemx.fr}
    }
    
\begin{document}
    	
    	\maketitle

    \begin{abstract}

    	Neural network classifiers (NNCs) are known to be vulnerable to malicious adversarial perturbations of inputs including those modifying a small fraction of the input features named sparse or $L_0$ attacks. Effective and fast $L_0$ attacks, such as the widely used Jacobian-based Saliency Map Attack (JSMA) are practical to fool NNCs but also to improve their robustness.
        In this paper, we show that penalising saliency maps of JSMA by the output probabilities and the input features of the NNC allows to obtain more powerful attack algorithms that better take into account each input's characteristics. This leads us to introduce improved versions of JSMA, named Weighted JSMA (WJSMA) and Taylor JSMA (TJSMA), and demonstrate through a variety of white-box and black-box experiments on three different datasets (MNIST, CIFAR-10 and GTSRB), that they are both significantly faster and more efficient than the original targeted and non-targeted versions of JSMA. 
        Experiments also demonstrate, in some cases, very competitive results of our attacks in comparison with the Carlini-Wagner (CW) $L_0$ attack, while remaining, like JSMA, significantly faster (WJSMA and TJSMA are more than 50 times faster than CW $L_0$ on CIFAR-10). Therefore, our new attacks provide good trade-offs between JSMA and CW for $L_0$ real-time adversarial testing on datasets such as the ones previously cited. 
    	
    \end{abstract}

    \section{Introduction}
    Deep learning classifiers are used in a wide variety of situations, such as vision, speech recognition, financial fraud detection, malware detection, autonomous driving, defense, and more.
    
    The ubiquity of deep learning algorithms in many applications, especially those that are critical such as autonomous driving \cite{DBLP:conf/cvpr/EykholtEF0RXPKS18,DBLP:journals/corr/abs-1802-06430} or that pertain to security and privacy \cite{DBLP:journals/corr/abs-1802-08908,DBLP:conf/ccs/SongSM19} makes their attack particularly useful. Indeed, this allows firstly to identify possible flaws in the intelligent model and secondly to set up a defense strategy to improve its reliability. 
    
    In this context, adversarial machine learning has appeared as a new branch that aims to thwart intelligent algorithms. Many techniques called adversarial attacks succeeded in fooling well-known architectures of neural networks, sometimes very astonishingly. Examples of these methods include for instance: Fast Gradient Sign Method \cite{FGM}, Basic Iterative Method  \cite{BIM}, Projected Gradient Descent \cite{PGD}, JSMA \cite{JSMA}, DeepFool \cite{DF}, Universal Adversarial Perturbations \cite{Moosavi-Dezfool} and CW attacks \cite{CW2}. 
    
    More formally, a neural network classifier (NNC) is an algorithm whose goal is to predict through a neural network which class an item $x$ belongs to, among a family of $K$ possible classes. It outputs a vector of probabilities $F(x)=(F_1(x),\cdots,F_K(x))$ where the label of $x$ is deduced by the rule $\text{label}(x)=\text{argmax}_k F_k(x)$.
    
    An adversarial example constructed on the item $x$, is an item $x^*$ specially crafted to be as close as possible to $x$ (with respect to some distance function), and such that it is classified by the NNC as $\text{label}(x^*) \neq \text{label}(x)$ (Non-Targeted (NT) attack), or even such that $\text{label}(x^*)=L$, with $L$ chosen by the attacker and such that $L\neq\text{label}(x)$ (Targeted attack).
    
    In this paper, we focus on the following class of attacks:
    
    \noindent\textbf{$L_0$ (or sparse) adversarial attacks.} They aim at generating adversarial samples while minimising the number of modified components. Sparse perturbations can be found in many real-life situations. As motivated in \cite{DBLP:conf/cvpr/ModasMF19}, \emph{sparse perturbations could correspond to some raindrops that reflect the sun on a``STOP" sign, but that are sufficient to fool an autonomous vehicle; or a crop-field with some sparse colorful flowers that force a UAV to spray pesticide on non affected areas.} This reveals very disturbing and astonishing properties of neural networks as it is possible to fool them by modifying few pixels \cite{DBLP:journals/tec/SuVS19,DBLP:conf/cvpr/ModasMF19}. Their study is therefore fundamental to mitigate their effects and take a step forward towards robustness of neural networks. The first proposed example of $L_0$ attacks is JSMA, a targeted attack \cite{JSMA}. In a computer vision application, JSMA achieved 97\% adversarial success rate by modifying on average 4.02\% input features per sample. \cite{JSMA} relates this result to the human capacity of visually detecting changes. An important quality of JSMA is that it is easy to understand, set up and it is relatively fast. For instance, relying on its $\mathtt{cleverhans}$ implementation \cite{Papernot2016TechnicalRO}, JSMA is able to generate 9 adversarial samples on the MNIST handwritten digits dataset \cite{lecun-mnisthandwrittendigit-2010} in only $2$ seconds on a laptop with $2$ CPU cores. Speed and also the ability to run adversarial attacks with limited resources are important criteria for real-life deployment of neural networks \cite{2019arXiv190707487E,2019arXiv190513399G,Lin2020OnTR}. JSMA obeys these constraints making its use widespread beyond computer vision applications such as in cybersecurity, anomaly detection and intrusion detection \cite{new_book,new_book1,articlex}. Later on, \cite{CW2} proposes the second example of targeted $L_0$ attacks known as CW $L_0$. The same paper shows that unlike JSMA, CW $L_0$ scales well to large datasets by considering the IMAGENET dataset \cite{imagenet_cvpr09}. CW $L_0$ is now the state-of-the-art $L_0$ targeted attack with the lower average $L_0$ distance. However, it is computationally-expensive, much slower than JSMA on small datasets (a factor of 20 times slower is reported in \cite{CW2}) and therefore, despite efficiency, it is less convenient for real-time applications. 
    
    Let us now briefly explain how the afore-mentioned attacks concretely work :
    
    \textbf{JSMA.} To fool NNCs, this attack relies on the Jacobian matrix of outputs with respect to inputs. By analysing this matrix, one can deduce how the output probabilities behave given a slight modification of an input feature. Consider a NNC $N$ as before and call  $Z(x)=(Z_1(x),\cdots,Z_K(x))$ the outputs of the second-to-last layer of $N$ (no longer probabilities, but related to the final output by a softmax layer). To generate an adversarial example from $x$, JSMA first computes the gradient $\nabla Z(x)$. The next step is to build a saliency map and find the most salient component $i$ that will then be changed:
    
    \begin{align}\label{eqq}
    S[x, t][i] = \left\{
    \begin{array}{ll}
    0 \;\;\; \textrm{if} \; \dfrac{\partial Z_t(x)}{\partial x_i} < 0 \;\; \textrm{or} \;\; \displaystyle \sum_{k \neq t} \dfrac{\partial Z_k(x)}{\partial x_i} > 0 \vspace{0.3cm}\\
    \dfrac{\partial Z_t(x)}{\partial x_i} \cdot \left\vert \displaystyle \sum_{k \neq t} \dfrac{\partial Z_k(x)}{\partial x_i} \right\vert \;\;\; \textrm{otherwise}.
    \end{array}
    \right.
    \end{align}
    $\dfrac{\partial Z_t(x)}{\partial x_i}$ and $\sum_{k \neq t} \dfrac{\partial Z_k(x)}{\partial x_i}$ in these maps quantify how much $Z_t(x)$ will increase and $\displaystyle \sum_{k \neq t} Z_k(x)$ will decrease, given a modification of the input feature $x_i$. Counting on the $Z_k$'s instead of the $F_k$'s has been justified in \cite{JSMA} by
    the extreme variations induced by the softmax layer. Then the algorithm selects the component: $i_{\text{max}}=\text{argmax}_i S[x, t][i]$ and increases $x_{i_{\text{max}}}$ by a default value $\theta$ before clipping to the valid domain. The same process is iterated until the class of $x$ is changed or a maximum allowed number of iterations is reached. This version of JSMA will be called one-component JSMA. A second, more effective, variant of JSMA recalled later relies on doubly indexed saliency maps. 
    
    \textbf{CW $L_0$ attack.} This method is obtained as a solution to the optimisation problem (assuming the domain of inputs is $[0,1]^n$):
    
    $$\text{Minimise}\ ||r||_0 + c f(x+r),\ \ x+r\in[0,1]^n$$ 
    
    where $||r||_0$ is the $L_0$ distance of the perturbation $r$ added to $x$. The recommended choice of $f$ is: $f(x)=(\text{max}_{i\ne t} Z_i(x)-Z_t(x))^+$. Since the $L_0$ distance is not convenient for gradient descent, the authors of \cite{CW2} solve this problem by making use of their $L_2$ attack and an algorithm that iteratively eliminates the components without much effect on the output classification. They finally obtain an effective $L_0$ attack that has a net advantage over JSMA. However, the main drawback of this attack is its high computational cost. 
    
    We summarise our main contributions as follows:
    
    \begin{itemize}[leftmargin=*,labelsep=5.8mm]
    	
    	\item For targeted misclassification, we introduce two variants of JSMA called Weighted JSMA (WJSMA) and Taylor JSMA (TJSMA). WJSMA applies a simple weighting to saliency maps by the output probabilities, and TJSMA does the same, while additionally penalising extremal input features. Both attacks are more efficient than JSMA according to several metrics (such as speed and mean $L_0$ distance). We present qualitative and quantitative results on MNIST and CIFAR-10 \cite{cifar10} supporting our claims. Moreover, although they are less efficient than CW $L_0$, our attacks have a major speed advantage over CW $L_0$  highlighted by measuring the execution time for each attack.  
    	
        \item For non-targeted (NT) misclassification, we improve the known NT variants of JSMA called NT-JSMA and Maximal-JSMA (M-JSMA) \cite{JSMA2}. We do this by introducing NT and M versions of WJSMA and TJSMA. Our attacks yield better results than NT-JSMA and M-JSMA. Also, they are as competitive but significantly much faster than NT CW $L_0$. These claims are illustrated through applications to attack a deep NNC on the GTSRB dataset \cite{Stallkamp2012} in the white/black-box modes.

    	\item We provide a deep comparison between WJSMA and TJSMA which is of independent interest. Our study concludes that in the targeted case, TJSMA is better than WJSMA. However, in the NT case, WJSMA is preferred over TJSMA, mainly due to the simplicity of its implementation. 	 	
    	
    	\item We provide fast and optimised implementations of the new attacks using TensorFlow and the $\mathtt{cleverhans}$ library \cite{Papernot2016TechnicalRO} that might help users working in adversarial machine learning (In all our experiments, we use the original implementation of JSMA available in the $\mathtt{cleverhans}$ library and the original code of CW publicly available. As for the NT versions of JSMA \cite{JSMA2}, whose implementations are not available, we re-implement these attacks using TensorFlow and $\mathtt{cleverhans}$. A link to the codes is provided at the end of subsection \ref{supp}) 
    
    \end{itemize}

    The rest of the paper is organised as follows. In Section \ref{sec3}, we discuss our main motivations and then introduce WJSMA and TJSMA as targeted attacks. Section \ref{sec4} is focused on NT and Maximal versions of WJSMA and TJSMA. Section \ref{experiments} is dedicated to several comparisons between our attacks, JSMA and CW $L_0$. We discuss attacking, defending with targeted/non-targeted attacks in both the white/black-box setups. Section \ref{conclusion} offers a conclusion and a summary of the main results of the paper. Finally, Appendices \ref{appendix1} and \ref{appendix2} are appendices dedicated to supplementary results and materials.

    \section{Targeted attacks}\label{sec3}
    The first attacks, presented in this section, are targeted and called Weighted JSMA (WJSMA) and Taylor JSMA (TJSMA). We give a detailed exposition of WJSMA and motivate the main idea leading to its derivation through a simple example. Then, we deduce TJSMA by applying once more and in a slightly different manner the same argument. In addition to the theoretical presentation, a few preliminary figures are given to illustrate a faster convergence of the new attacks in comparison with JSMA.   
    \subsection{Weighted Jacobian-based Saliency Map Attack (WJSMA)}\label{weighted}
    
    The main idea here is to penalise gradients associated with small probabilities so as to mitigate their influences in saliency maps. The goal is to obtain more balanced saliency maps than those proposed by JSMA. We first give a concrete example illustrating a concrete limitation of JSMA which motivated this work.
    
    \textbf{Motivating example.} Assume a number of classes $K\ge 4$ and for some input $x$: $F_1(x)=0.5$, $F_2(x)=0.49$, $F_3(x)=0.01$ and $F_k(x)=0$ for all $4\le k\leq K$. Consider the problem of generating an adversarial sample to $x$ with label $t=2$. In order to decrease $\displaystyle \sum_{k \neq 2} Z_k(x)$, the first iteration of JSMA relies on the gradients $\nabla Z_k(x), k\ne 2$. Our main observation is that since the probabilities $F_k(x)=0$ for $4\le k\leq K$ are already in their minimal values, taking into account $\nabla Z_k(x)$ for these values of $k$ in the search of  $i_{\text{max}}$ is unnecessary. In other words, by only acting on gradients, JSMA does not consider the crucial constraints on probabilities: $F_k(x)\ge 0$. Moreover, instead of relying equally on $\nabla Z_1(x)$ and  $\nabla Z_3(x)$, for this example one would ``bet" on $\nabla Z_1(x)$ than $\nabla Z_3(x)$ as the possible decrease for $F_1(x)$ is high (up to $0.5$) and $F_3(x)$ is relatively small, thus hard to decrease further.
    
    \textbf{Weighted JSMA (WJSMA).} Our first solution to the previous issue is WJSMA. Its principle is to penalise each gradient $\dfrac{\partial Z_k(x)}{\partial x_i}$, where $k\neq t$, by the probability $F_k(x)$. Besides the intuition of this idea, we will provide a justification of it by a classical log softmax argument. First, we compute:
    
    $$\dfrac{\partial}{\partial x_i}\log F_t(x) = (1-F_t(x))\dfrac{\partial Z_t}{\partial x_i}(x)-\sum_{k\ne t} F_k(x) \dfrac{\partial Z_k}{\partial x_i}(x)$$
    
    One way to maximise this derivative with respect to $i$, is to maximise $A=\dfrac{\partial Z_t}{\partial x_i}(x)$ and minimise $B=\sum_{k\ne t} F_k(x) \dfrac{\partial Z_k}{\partial x_i}(x)$ under the constraints $A>0$ and $B < 0$. These constraints ensure, in particular, that $\dfrac{\partial F_t}{\partial x_i}(x)$ remains positive, a fact which is not necessarily guaranteed under JSMA constraints. According to this, we introduce one-component weighted saliency maps as follows:

    \begin{align}
    S^W[x, t][i] = \left\{
    \begin{array}{ll}
    0 \;\;\; \textrm{if} \; \dfrac{\partial Z_t(x)}{\partial x_i} < 0 \;\; \textrm{or} \;\; \displaystyle \sum_{k \neq t} F_k(x)\dfrac{\partial Z_k(x)}{\partial x_i} > 0 \vspace{0.3cm}\\
    \dfrac{\partial Z_t(x)}{\partial x_i} \cdot \left\vert \displaystyle \sum_{k \neq t} F_k(x)\dfrac{\partial Z_k(x)}{\partial x_i} \right\vert \;\;\; \textrm{otherwise}.\nonumber\
    \end{array}
    \right.
    \end{align}
    
    Based on these maps, we present Algorithm \ref{algo11}, the first version of WJSMA, to generate targeted adversarial samples.
    
    \begin{algorithm}[!h]
    	\caption{Generating adversarial samples by WJSMA: version 1}\label{algo11}
    	\textbf{Inputs:}\hspace*{0.1cm}$N$: a NNC, $Z$: second-to-last output of $N$, $x$: input to $N$, $t$: target label ($t\ne \text{class}(x)$), $\mathtt{maxIter}$: maximum number of iterations, $\theta_{\text{min}}, \theta_{\text{max}}$ lower and upper bounds for features values, $\theta$: positive default increase value. \\
    	\textbf{Output:}\hspace*{0.1cm}$x^*$: adversarial sample to $x$. \\
    	\hrule
    	
    \begin{algorithmic}
          
            \STATE $x^* \leftarrow x$
            \vspace{0.1cm}
            \STATE $\mathtt{iter} \leftarrow 0$
            \vspace{0.1cm}
            \STATE $\Gamma \leftarrow \llbracket 1,|x| \rrbracket \setminus \{ p \in \llbracket 1,|x| \rrbracket \: | \: x[p] = \theta_{max} \}$
            \vspace{0.2cm}
            \WHILE{$\text{class}(x^*) \neq t \textbf{ and }\mathtt{iter}< \mathtt{maxIter} \textbf{ and } \Gamma \neq \ \emptyset$}
            \vspace{0.1cm}
            \STATE $p_{max} = \argmax_{p\in \Gamma} S^W[x^*, t](p)$
            \vspace{0.2cm}
            \STATE Modify $x^*$ by $x^*[p_{max}]=\mathtt{Clip}_{[\theta_{\text{min}},\theta_{\text{max}}]}(x^*[p_{max}]+\theta)\ \ // \mathtt{Clip}\ \ \text{is the clipping function}$
            \vspace{0.2cm}
            \STATE Remove $p_{\text{max}}$ from $\Gamma$
            \vspace{0.2cm}
            \STATE $\mathtt{iter}++$
            \vspace{0.2cm}
            \ENDWHILE
            \vspace{0.1cm}
            \STATE \textbf{return} $x^{*}$
    
    \end{algorithmic}
        
    \end{algorithm}
    
    When the output $x^*$ of Algorithm \ref{algo11} satisfies $\text{class}(x^*)=t$, the attack is considered as successful.
    
    
    To relax a bit the search of salient components and motivated by a computer vision application, \cite{JSMA} introduces saliency maps indexed by pairs of components. The main argument is that the conditions required in (\ref{eqq}) may be too strict for some applications and very few components will verify it. Our doubly indexed versions of these maps are introduced in the same way as follows:
    \begin{align}
    S^W[x, t][i,j]= \left\{
    \begin{array}{ll}
    0 \;\;\; \textrm{if} \; \displaystyle \sum_{a \in \{i,j\}}\dfrac{\partial Z_t(x)}{\partial x_{a}} < 0 \;\; \textrm{or} \;\; \displaystyle \sum_{k \neq t} F_k(x)\sum_{a \in \{i,j\}}\dfrac{\partial Z_k(x)}{\partial x_{a}}  > 0 \vspace{0.3cm}\\
    \displaystyle\sum_{a \in \{i,j\}}\dfrac{\partial Z_t(x)}{\partial x_{a}} \cdot \left\vert \displaystyle \sum_{k \neq t} F_k(x)\sum_{a \in \{i,j\}}\dfrac{\partial Z_k(x)}{\partial x_{a}} \right\vert\ \textrm{otherwise}.
    \end{array}
    \right.
    \end{align}
    
    Algorithm \ref{algo12} presented below relies on $S^W[x, t][i,j]$ to generate targeted adversarial samples and is our second version of WJSMA.
    
    \begin{algorithm}[!h]
    	\caption{Generating adversarial samples by WJSMA: version 2} \label{algo12}
    	\hspace*{0.1cm} Same inputs and output as Algorithm \ref{algo11}.\\
    	\hrule
    \begin{algorithmic}\label{crafting0}
          
            \STATE $x^* \leftarrow x$
            \vspace{0.1cm}
            \STATE $\mathtt{iter} \leftarrow 0$
            \vspace{0.1cm}
            \STATE $\Gamma \leftarrow \{(p,q), p, q \in \llbracket 1,|x| \rrbracket, x[p] \neq \theta_{max}, x[q] \neq \theta_{max}\}$
            \vspace{0.1cm}
            
            \WHILE{$\text{class}(x^*) \neq t \textbf{ and }\mathtt{iter}< \mathtt{maxIter} \textbf{ and } \Gamma \neq \ \emptyset$}
            \vspace{0.2cm}
            \STATE $(p_{\text{max}}, q_{\text{max}}) = \text{argmax}_{p,q\in \Gamma} S^W[x^*, t](p,q)$
            \vspace{0.2cm}
            \STATE Modify $x^*$ by $x^*[a]=\mathtt{Clip}_{[\theta_{\text{min}},\theta_{\text{max}}]}(x^*[a]+\theta),\ a=p_{\text{max}}, q_{\text{max}}$
            \vspace{0.2cm}
            \STATE Remove $(p_{\text{max}}, q_{\text{max}})$ from $\Gamma$
            \vspace{0.2cm}
            \STATE $\mathtt{iter}++$
            \vspace{0.2cm}
            \ENDWHILE
            \vspace{0.2cm}
            \STATE \textbf{return} $x^{*}$
    \end{algorithmic}
        
    \end{algorithm}
    In practice and despite the fact that each iteration of Algorithm \ref{crafting0} is more computationally-expensive than each iteration of Algorithm \ref{algo11}, we find that it gives better results. This agrees with the recommendations of \cite{JSMA} on the superiority of two-components versions for JSMA. Finally, we notice that while in the two previous algorithms, the selected components are always augmented by positive default values,  decreasing versions can be given following a similar logic.
    
    \subsection{Taylor Jacobian-based Saliency Map Attack (TJSMA)}

    The principle of our second attack, TJSMA, is to additionally penalise the choice of feature components that are close to the maximum value $\theta_{\text{max}}$. Assume $i$ and $j$ have the same WJSMA score $S^W[x,t][i]=S^W[x,t][j]$ and that $x_i$ is very close to $\theta_{\text{max}}$, while $x_j$ is far enough from $\theta_{\text{max}}$. In this case, looking for more impact, TJSMA prefers $x_j$ over $x_i$. Concretely, we simultaneously maximise $S_1=\theta_{\text{max}}-x_i$ and $S_2=\dfrac{\partial}{\partial x_i}\log p_t(x)$ by maximising the product $S=S_1 S_2$. Accordingly, we introduce new saliency maps for one and two-components selection as follows: 
    \begin{align}\label{ss}
    S^T[x, t][i]= \left\{
    \begin{array}{ll}
    0 \;\;\; \textrm{if} \;  \alpha_{i}< 0 \;\; \textrm{or} \;\; \beta_{i} > 0 \vspace{0.3cm}\\
    \alpha_{i} |\beta_{i}|\ \ \textrm{otherwise}.
    \end{array}
    \right.
    \end{align}
    
    where 
    
    $$\alpha_{i}=(\theta_{\text{max}}-x_i)\dfrac{\partial Z_t(x)}{\partial x_{i}},\ \ \beta_{i}= (\theta_{\text{max}}-x_i) \sum_{k \neq t}  F_k(x)\dfrac{\partial Z_k(x)}{\partial x_{i}}$$
    
    and 
    \begin{align}\label{ss}
    S^T[x, t][i,j]= \left\{
    \begin{array}{ll}
    0 \;\;\; \textrm{if} \;  \alpha_{i,j}< 0 \;\; \textrm{or} \;\; \beta_{i,j} > 0 \vspace{0.3cm}\\
    \alpha_{i,j} |\beta_{i,j}|\ \ \textrm{otherwise}.
    \end{array}
    \right.
    \end{align}
    
    where 
    
    $$\alpha_{i,j}=\sum_{a \in \{i,j\}}(\theta_{\text{max}}-x_a)\dfrac{\partial Z_t(x)}{\partial x_{a}},\ \ \beta_{i,j}=\sum_{k \neq t}\sum_{a \in \{i,j\}}   F_k(x)(\theta_{\text{max}}-x_a)\dfrac{\partial Z_k(x)}{\partial x_{a}}$$
    
    Due to the presence of the Taylor terms $(\theta_{\text{max}}-x_a)\dfrac{\partial Z_k(x)}{\partial x_{a}}$, we call these maps Taylor saliency maps. We introduce one and two-components TJSMA following exactly Algorithms \ref{algo11}, \ref{algo12} and only replacing $S^W$ with $S^T$.
    
    \begin{figure}[H]
        \centering
        \begin{subfigure}[b]{0.48\textwidth}
            \includegraphics[width=\textwidth]{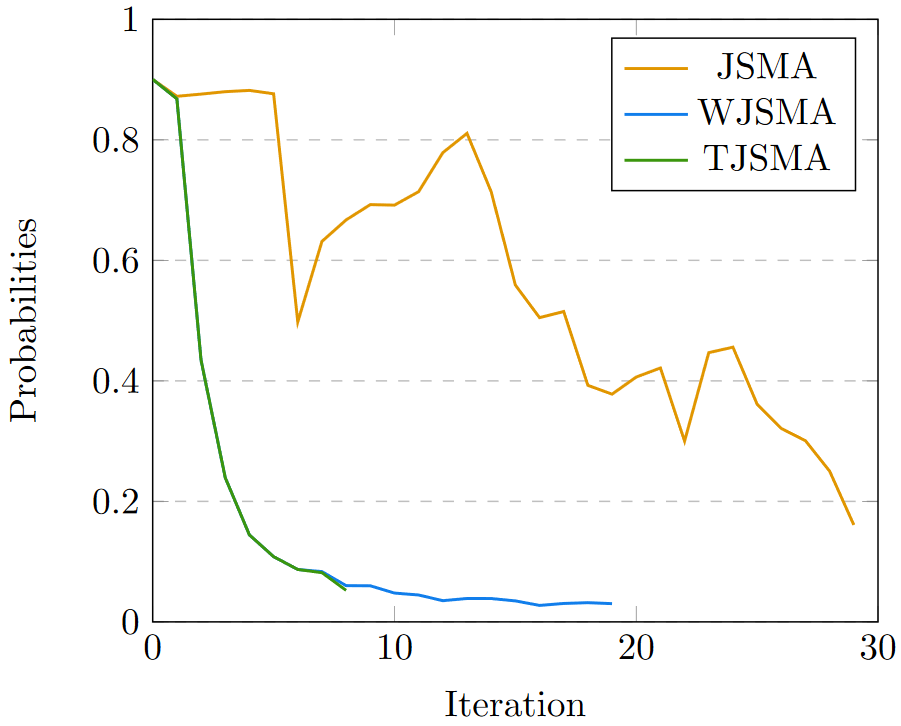}
            \caption{Origin class}
            \label{probabilities_a}
        \end{subfigure}\hspace{2mm}
        \begin{subfigure}[b]{0.48\textwidth}
            \includegraphics[width=\textwidth]{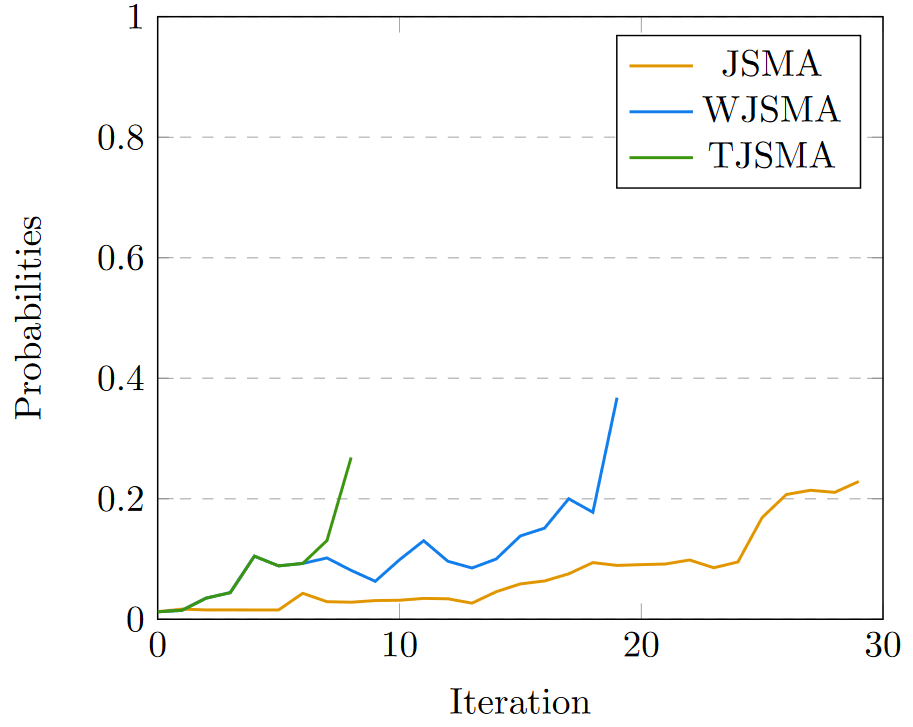}
            \caption{Target class}
            \label{probabilities_b}
        \end{subfigure}
        \caption{Evolution of the origin and target class  probabilities until the target class is reached for JSMA, WJSMA and TJSMA changing the image of a one into a five.}
        \label{probabilities}
    \end{figure}
    
    Figures \ref{probabilities_a} and \ref{probabilities_b} offer a concrete illustration of the convergence of JSMA, WJSMA and TJSMA. In particular, we observe that WJSMA and TJSMA decrease/increase the predicted/targeted probability of the original/targeted class much sooner than JSMA. Also, we note that TJSMA behaves like WJSMA until it is able to find a more vulnerable component that makes it converge much faster.
    
    \section{Non-targeted attacks}\label{sec4}
    NT variants of JSMA have been studied in \cite{JSMA2}. In particular, the paper introduces NT-JSMA-F, NT-JSMA-Z based on NT saliency maps whose role is to select the most salient pairs of components to decrease as much as possible the probability of the current class. The notations -F and -Z indicate if the saliency maps either use the $F_k$'s or the $Z_k$'s. Second, \cite{JSMA2} proposes maximal JSMA (M-JSMA) as a more flexible attack allowing both increasing/decreasing features and also combining targeted/non-targeted strategies at the same time.
    
    In what follows, we again leverage the idea of penalising saliency maps to give our proper NT JSMA attacks. For a unified presentation, we use the letter $X$ to denote either $W$ (Weighted) or $T$ (Taylor) and the letter $Y$ to denote either $Z$ (logits) or $F$ (probabilities). We notice that while the first version of JSMA uses the logits, variants that rely on the $F_k$'s also demonstrated good performances \cite{CW2,JSMA2}. Thus for a more complete study, we give versions with both $Z$ and $F$.

    By a NT reasoning, similar to the previous section, we define Weighted and Taylor NT saliency maps as follows: 
    
    \begin{align}\label{ss}
    S^{X,Y}[x, t][i,j]= \left\{
    \begin{array}{ll}
    0 \;\;\; \textrm{if} \;  \alpha_{i,j}^{X,Y}> 0 \;\; \textrm{or} \;\; \beta^{X,Y}_{i,j} < 0 \vspace{0.3cm}\\
    |\alpha^{X,Y}_{i,j}| \beta^{X,Y}_{i,j}\ \ \textrm{otherwise}.
    \end{array}
    \right.
    \end{align}
    
    where, for $X=W$,
    
    $$\alpha^{W,Y}_{i,j}=\sum_{a \in \{i,j\}}\dfrac{\partial Y_t(x)}{\partial x_{a}},\ \ \beta^{W,Y}_{i,j}=\sum_{k \neq t}\sum_{a \in \{i,j\}}   F_k(x)\dfrac{\partial Y_k(x)}{\partial x_{a}}$$
    and, for $X=T$,
    
    $$\alpha^{T,Y}_{i,j}=\sum_{a \in \{i,j\}}(\theta_{\text{max}}-x_a)\dfrac{\partial Y_t(x)}{\partial x_{a}},\ \ \beta^{T,Y}_{i,j}=\sum_{k \neq t}\sum_{a \in \{i,j\}}   F_k(x)(\theta_{\text{max}}-x_a)\dfrac{\partial Y_k(x)}{\partial x_{a}}$$
    
    These maps can be motivated, like in the previous section, by considering the simple one-component case: $i=j$. For example, the role of the penalisation by $F_k(x)$ in $S^{W,Z}$ is to reduce the impact of high gradients $\dfrac{\partial Z_k(x)}{\partial x_{i}}$ when the probability $F_k(x)$ is very small (in this case we penalise choosing $k$ as a new target for $x$).
    
    
    Relying on saliency maps, Algorithm \ref{algo28} below presents our improvements of NT-JSMA-Z and NT-JSMA-F. For the sake of simplification, we have employed a unified notation NT-XJSMA-Y. For example, when we use $S^{W,Z}$, the obtained attack is NT-WJSMA-Z. Again, we only write the increasing version.
    
    \begin{algorithm}[H]
    	\caption{NT-WJSMA and NT-TJSMA attacks} \label{algo28}
    	\textbf{Inputs:}\hspace*{0.1cm}$x$: input to $N$ with label $t$, $\text{maxIter}$: maximum number of iterations, X $\in \{ W, T \}$, Y $\in \{ Z, F \}$\\
    	\textbf{Output:}\hspace*{0.1cm}$x^*$: adversarial sample to $x$.
    	\hrule
    	\begin{algorithmic}\label{crafting}
    		\STATE $x^* \leftarrow x$
    		\STATE $\mathtt{iter} \leftarrow 0$
    		\STATE $\Gamma \leftarrow \{(p,q), p, q \in \llbracket \theta_{\text{max}},|x| \rrbracket, x[p] \neq \theta_{\text{max}}, x[q] \neq \theta_{\text{max}}\}$
    		\WHILE{$\text{class}(x^*) \neq t \textbf{ and }\mathtt{iter}< \mathtt{maxIter} \textbf{ and } |\Gamma| \geq 2$}
    		\STATE $(p_{\text{max}}, q_{\text{max}}) = \text{argmax}_{p,q\in \Gamma} S^{X,Y}[x^*, t](p,q)$
    		\STATE Modify $x^*$ by $x^*[p_{\text{max}}], x^*[q_{\text{max}}]=\theta_{\text{max}}$
    		\STATE Remove $(p_{\text{max}}, q_{\text{max}})$ from $\Gamma$
    		\STATE $\mathtt{iter}++$
    		\ENDWHILE
    		\STATE \textbf{return} $x^{*}$
    	\end{algorithmic}
    \end{algorithm}
    We now turn to extensions of M-JSMA \cite{JSMA2}. This attack modifies the pairs of components achieving the best score among NT-JSMA and all possible targeted (including increasing and decreasing) JSMA. It has a greater capacity to craft adversarial samples but is relatively slower than NT-JSMA. Our extensions of M-JSMA, which we call M-WJSMA and M-TJSMA, are described in Algorithm \ref{algo299}.
    \begin{algorithm}[H]
    	\caption{M-WJSMA and M-TJSMA attack} \label{algo299}
    	\textbf{Inputs:}\hspace*{0.1cm}$x$: input to $N$ with label $t$, $\text{maxIter}$: maximum number of iterations, X $\in \{ W, T \}$, Y $\in \{ Z, F \}$\\
    	\textbf{Output:}\hspace*{0.1cm}$x^*$: adversarial sample to $x$.
    	\hrule
    	\begin{algorithmic}\label{crafting}
    		\STATE $x^* \leftarrow x$
    		\STATE $\mathtt{iter} \leftarrow 0$
    		\STATE $\Gamma \leftarrow \{(p,q), p, q \in \llbracket 1,|x| \rrbracket, x[p], x[q] \neq \theta_{\text{min}}, \theta_{\text{max}}\}$
    		\vspace{0.1cm}
    		\WHILE{$\text{class}(x^*) \neq t \textbf{ and }\mathtt{iter}< \mathtt{maxIter} \textbf{ and } |\Gamma| \geq 2$}
    		\STATE - Compute all targeted increasing/decreasing saliency maps scores $S^{X,Y}[x,s](p,q), s\neq t$ (Section \ref{sec3}) and all NT increasing/decreasing saliency maps scores $S^{X,Y}[x,t](p,q)$ (\ref{ss}).
    		\vspace{0.1cm}
    		\STATE - Choose $(p_{\text{max}}, q_{\text{max}})$ achieving the best score and saturate $x^*[p_{\text{max}}], x^*[q_{\text{max}}]$ to $\theta_{\text{min} }$ or $\theta_{\text{max} }$ according to the chosen saliency map.
    		\vspace{0.1cm}
    		\STATE - Remove $(p_{\text{max}}, q_{\text{max}})$ from $\Gamma$
    		\vspace{0.1cm}
    		\STATE- $\mathtt{iter}++$
    		\vspace{0.1cm}
    		\ENDWHILE
    		\vspace{0.1cm}
    		\STATE \textbf{return} $x^{*}$
    	\end{algorithmic}
    	
    \end{algorithm}
    
    Note that saliency maps $S^{W,Y}$ for targeted or non-targeted, features-increasing or features-decreasing attacks are exactly the same (one only needs to decide between an argmax or argmin).  This is not the case for M-TJSMA since for example $(\theta_{\text{max} }-x_a)$ has to be changed to $x_a-\theta_{\text{min} }$ when decreasing features. As a consequence of this fact, M-WJSMA is less cumbersome to implement than M-TJSMA. Trying to keep the paper and code as simple as possible, we choose M-WJSMA over M-TJSMA and do not include M-TJSMA in our experiments. M-WJSMA already gives us satisfactory results. 
    
    \section{Experiments}\label{experiments}
    
    This section is dedicated to a variety of experiments on the proposed attacks and several comparisons with the state-of-the-art methods. The first part focuses on targeted attacks and provides intensive comparisons between JSMA, WJSMA and TJSMA on deep NNCs on MNIST and CIFAR-10 as well as comparisons with CW $L_0$. In the second part, we show that our approach is still  relevant for non-targeted $L_0$ misclassification in both the white-box and black-box modes. A particular emphasis will be put on the speed of our attacks in comparison with CW $L_0$.
    
    \subsection{Experiments on targeted attacks}
    
    In the following, we give attack and defense applications illustrating the interest of WJSMA and TJSMA over JSMA. In doing so, we also compare WJSMA and TJSMA and report overall better results for TJSMA despite the fact that for a large part of samples WJSMA outperforms TJSMA. Finally, we provide a comparison with CW $L_0$ attack and comment on all the obtained results. 
    
    The datasets used in this section are: 
    
    \textbf{MNIST \cite{lecun-mnisthandwrittendigit-2010}.} This dataset contains 70,000 $28\times 28$ greyscale images in $10$ classes, divided into 60,000 training images and 10,000 test images. The possible classes are digits from $0$ to $9$. 
    
    
    \textbf{CIFAR-10 \cite{cifar10}.} This dataset contains 60,000 $32\times 32\times 3$ RGB images. There are 50,000 training images and 10,000 test images. These images are divided into $10$ different classes (airplane, automobile, bird, cat, deer, dog, frog, horse, ship, truck), with 6,000 images per class. 
    
\begin{figure}[H]
        \centering
        \includegraphics[width=0.7\textwidth]{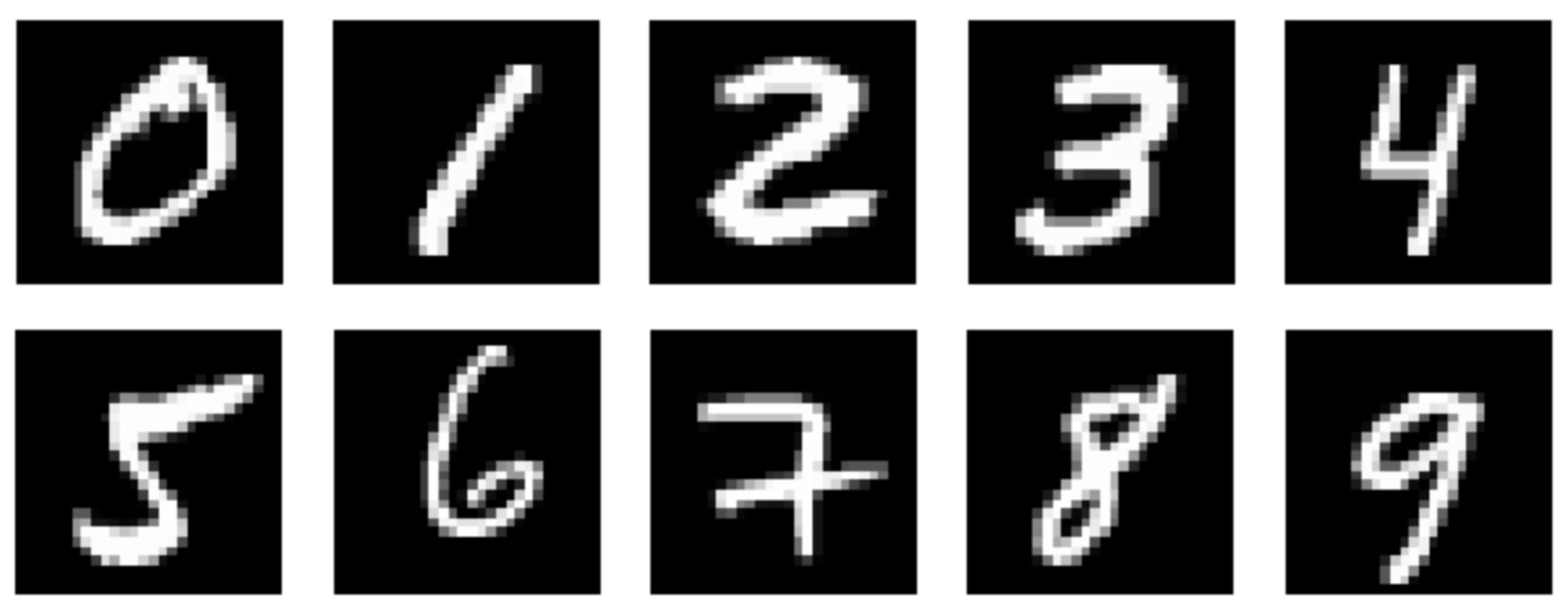}
        \caption{Examples of images from MNIST}
        \label{mnistimages}
    \end{figure}

\begin{figure}[H]
        \centering
        \includegraphics[width=0.7\textwidth]{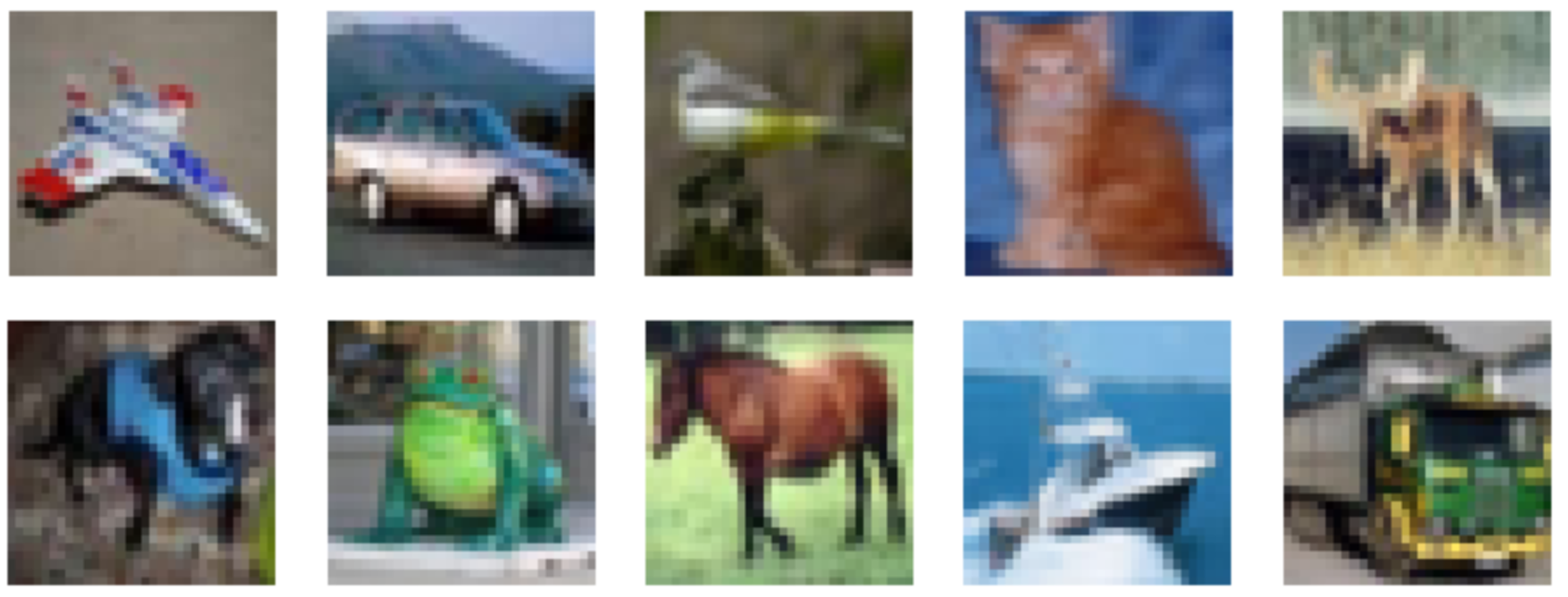}
        \caption{Examples of images from CIFAR-10}
        \label{cifar10images}
    \end{figure}

    Figures \ref{mnistimages} and \ref{cifar10images} display one sample per class, from MNIST and CIFAR-10 respectively. 
    
    On each dataset, a deep NNC is trained and then confronted to attacks.
    
    \textbf{NNC on MNIST.} Similarly to \cite{JSMA}, we consider LeNet-5 model on this dataset \cite{Lecun98gradient-basedlearning}. Its full architecture is described in Appendix \ref{appendix1}. We implement and train this model using a $\mathtt{cleverhans}$ model that optimises crafting adversarial examples. The number of epochs is fixed to 20, the batch-size to 128, the learning rate to 0.001 and the Adam optimizer is used.  Training results in 99.98\% accuracy on the training dataset and 99.49\% accuracy on the test dataset.
    
    \textbf{NNC on CIFAR-10.} We consider a more complex NNC, trained to reach a good performance on this dataset. Its architecture is inspired by the All Convolutional model proposed in $\mathtt{cleverhans}$ and is fully described in Appendix \ref{appendix1}. Likewise, this model is implemented and trained using $\mathtt{cleverhans}$ for 10 epochs, with a batch size of 128, a learning rate of 0.001 and the Adam optimizer. Training results in a 99.96\% accuracy on the training dataset and 83.81\% accuracy on the test dataset.
    \newline
    
    Our first objective is to compare the performances between JSMA, WJSMA and TJSMA on the previous two NNCs. To run JSMA, we use its original implementation, available in $\mathtt{cleverhans}$. We have also adapted the code to WJSMA and TJSMA thus obtaining fast implementations of these two attacks.

    For testing, we consider all images in MNIST, the first 10,000 training and 10,000 test images of CIFAR-10. Moreover, we only test on the images which are correctly predicted by the neural networks as this makes more sense. In this way, the attacks are applied to the whole training set and the 9,949 well-predicted images of the MNIST test images. Similarly, CIFAR-10 adversarial examples are crafted from the well-predicted 9,995 images of the first training 10,000 images and the 8,381 well-predicted test images.
    
    To compare the three attacks, we rely on the notion of maximum distortion of adversarial samples defined as the ratio of altered components to the total number of components.
    Following \cite{JSMA}, we choose a maximum distortion of $\gamma=14.5\%$ on the adversarial samples from MNIST, corresponding to $\mathtt{maxIter}=\lfloor \frac{784*\gamma}{2*100}\rfloor$. On CIFAR-10, we fix $\gamma=3.7\%$ in order to have the same maximum number of iterations for both experiments. This allows a comparison between the attacks in two different settings. Furthermore, for both experiments, we set $\theta=1$ (note that $\theta_{\text{min}}=0,\ \theta_{\text{max}}=1$).
    \newline
    
    We report the metrics: 
    
    \begin{enumerate}[leftmargin=*,labelsep=4.9mm]
        \item  Success rate: This is the percentage of successful adversarial examples, i.e crafted before reaching the maximal number of iterations $\mathtt{maxIter}$,
        
        \item Mean $L_0$ distance: This is the average number of altered components of the successful adversarial examples,
        
        \item Strict dominance of an attack: Percentage of adversarial examples for which this attack does strictly fewer iterations than the two other ones (As additional results, we give in Appendix \ref{appendix2} more statistics on the dominance between any two attacks.),
        
        \item Run-time of an attack on a set of samples targeting every possible class. 
        
    \end{enumerate}
    
    Results on the metrics 1 and 2 are shown in Table \ref{mnist} for MNIST and Table \ref{cifar10res} for CIFAR-10.
    \begin{table}
    	\centering
    	\caption{Comparison between JSMA, WJSMA and TJSMA on MNIST.}
    	\label{mnist}
    	\begin{tabular}{C{6cm} C{2cm} C{2cm} C{2cm} }
    	    \toprule
    		\textbf{Metric} & \textbf{JSMA} & \textbf{WJSMA} & \textbf{TJSMA}\\
    		\midrule
    		\multicolumn{4}{c}{\textbf{Targeted (Training dataset: Nb of well predicted images=60,000)}} \\
    		\midrule
    	    Success rate&87.68\%&\textbf{97.14\%}&\textbf{98.66\%}\\
    	    \midrule
    		Mean $L_0$ distance on successful samples&44.34&\textbf{37.86}&\textbf{35.22}\\
    		\midrule
    		\multicolumn{4}{c}{\textbf{Targeted (Test dataset: Nb of well predicted images=9,949))}} \\
    		\midrule
    		Success rate&87.34\%&\textbf{96.98\%}&\textbf{98.68\%}\\
    		\midrule
    		Mean $L_0$ distance on successful samples&44.63&\textbf{38.10}&\textbf{35.50}\\
    		\bottomrule
    	\end{tabular}
    \end{table}
    
    \begin{table}
    	\centering
    	\caption{Comparison between JSMA, WJSMA and TJSMA on CIFAR-10.}
    	\label{cifar10res}
    	\begin{tabular}{C{6cm}C{2cm}C{2cm}C{2cm}}
    		\toprule
            \textbf{Metric} & \textbf{JSMA} & \textbf{WJSMA} & \textbf{TJSMA}\\			
            \midrule
    		\multicolumn{4}{c}{\textbf{Targeted (Training dataset: Nb of well predicted images=9 995)}} \\
    	  \midrule
    	  Success rate&86.17&\textbf{95.91\%}&\textbf{97.40\%}\\
    	  Mean $L_0$ distance on successful samples&47&\textbf{38.54}&\textbf{36.86}\\
    	  \midrule
    	  
    		\multicolumn{4}{c}{\textbf{Targeted (Test dataset: Nb of well predicted images=8 381))}} \\
    	\midrule
    	Success rate &84.91&\textbf{94.99\%}&\textbf{96.96\%}\\
    	Mean $L_0$ distance on successful samples &46.13&\textbf{38.82}&\textbf{37.45}\\
    	\bottomrule
    	\end{tabular}
    \end{table}
    
    First, we observe that overall, WJSMA and TJSMA significantly outperform JSMA according to metrics 1 and 2. Here are more comments: 
    
    \textbf{On MNIST.} Results in terms of success rate are quite remarkable for WJSMA and TJSMA respectively outperforming JSMA with near $9.46, 10.98$ percentage points (pp) on the training set and $9.46, 11.34$ pp on the test set. The gain in the average number of altered components exceeds 6 components for WJSMA and 9 components for TJSMA in both experiments. 
    
    \textbf{On CIFAR-10.} WJSMA and TJSMA outperform JSMA in success rate by near $9.74, 11.23$ pp on the training set and more than $10, 12$ pp on the test set. For both training and test sets, we report better mean $L_0$ distances exceeding $7$ features in all cases and up to $10.14$ features for TJSMA on the training set. 
     
\begin{figure}[H]
    	\centering
    	\begin{subfigure}[b]{0.46\textwidth}
    	    \includegraphics[width=\textwidth]{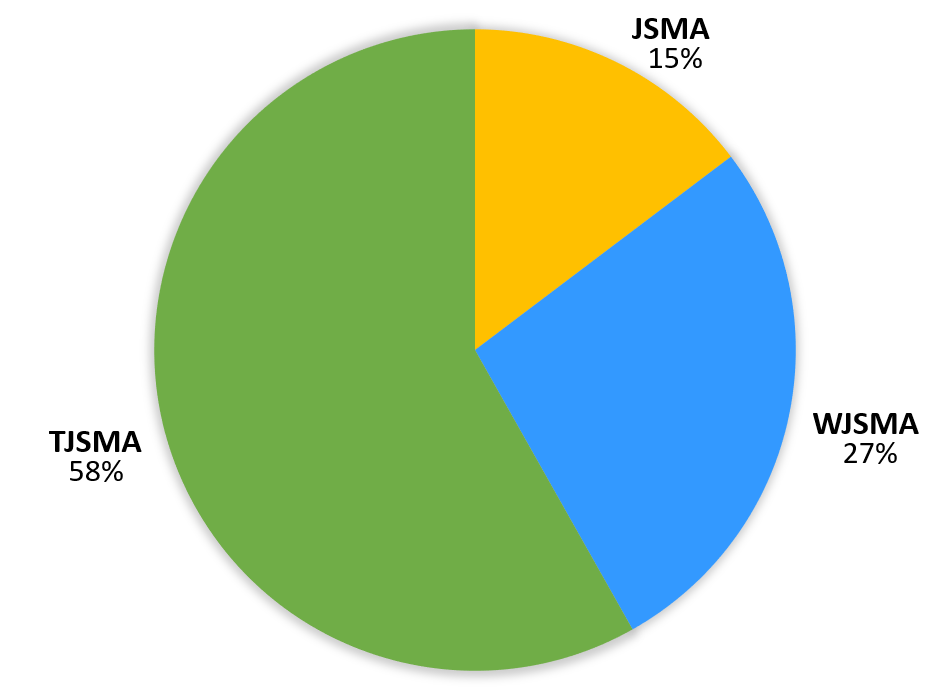}
    	    \caption{MNIST}
    	    \label{dominance_mnist}
    	\end{subfigure}
    	\begin{subfigure}[b]{0.46\textwidth}
    	    \includegraphics[width=\textwidth]{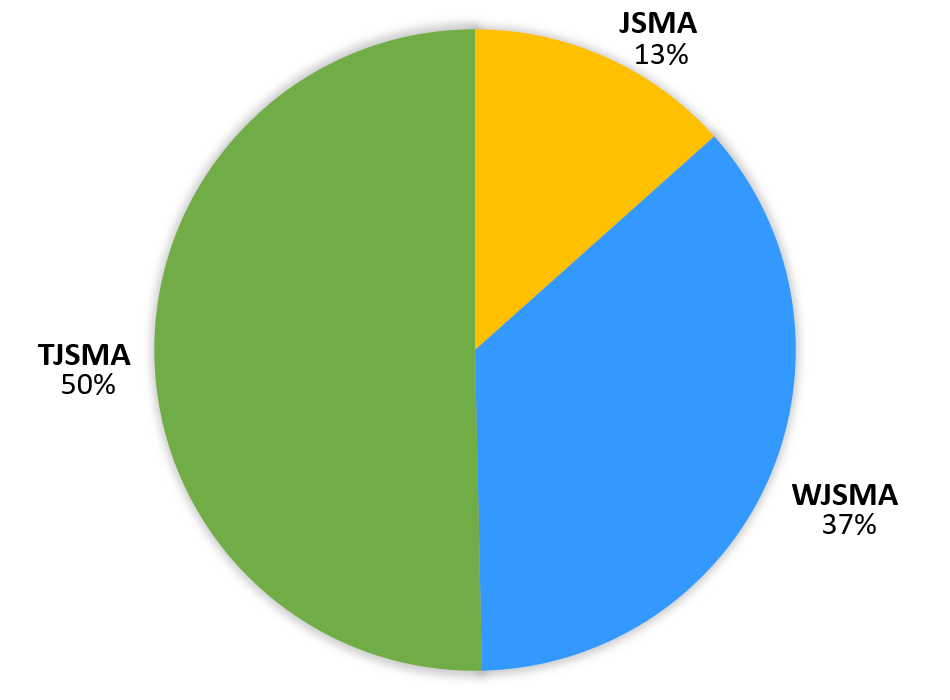}
    	    \caption{CIFAR-10}
    	    \label{dominance_cifar10}
    	\end{subfigure}
    	\caption{Distribution of the (strict) dominance of  JSMA, WJSMA and TJSMA over the MNIST (\ref{dominance_mnist}) and CIFAR10 (\ref{dominance_cifar10}) datasets (training and test sets included)}
    	\label{dominances}
    \end{figure}

    \noindent\textbf{Dominance of the attacks.} Figure \ref{dominances} illustrates the (strict) dominance of the attacks for the two experiments. In these statistics, we do not count the samples for which TJSMA and WJSMA have the same number of iterations and strictly less than JSMA.    
    
    For both experiments, TJSMA has a notable advantage over WJSMA and JSMA. The benefit of WJSMA over JSMA is also considerable. This shows that, in most cases, WJSMA and TJSMA craft better adversarial examples than JSMA, while being faster. Our results are indeed better when directly comparing WJSMA or TJSMA with JSMA. As additional results, we give in the appendix the statistics for the pairwise dominance between the attacks. As it might be expected, both WJSMA and TJSMA dominate JSMA, moreover TJSMA dominate WJSMA.
    
    \subsubsection{Avoid confusion.}\label{confusion} It is important to stress that our results do not contradict \cite{JSMA} obtaining $97\%$ success rate on LeNet-5. Indeed, we use a more efficient LeNet-5 model (the one in \cite{JSMA} has $98.93\%$ and $99.41\%$ accuracies on the training and test sets). For completeness, we also generated a second model (with $99.34\%$ and $98.94\%$ accuracies on the training and test sets) and evaluated the three attacks on the first $1,000$ test MNIST images. We obtain $96.7\%$ success rate for JSMA (very similar to \cite{JSMA}) and more than $99.5\%$ for WJSMA and TJSMA. We preferred to work with the more effective model as this makes the paper shorter and moreover, it values more our approach (giving us more advantage with respect to JSMA).
    
    \subsubsection{Run-time comparison.} In order to have a meaningful speed comparison between the three attacks, we computed the time needed for each attack to successfully craft the first 1,000 test MNIST images in the targeted mode. Results are shown in Table \ref{tab3} and reveal that TJSMA/WJSMA are 1.41/1.28 times faster than JSMA. These performances were measured on a machine equipped with an Intel Xeon 6126 processor and a Nvidia Tesla P100 graphics processor. Note that for WJSMA/TJSMA, the additional computations of one iteration compared to JSMA are negligible (simple multiplications). Thus the difference in speed between the attacks is mainly due to 
    the number of iterations for each attack.
    
        \begin{table}[]
            \centering
            \caption{Time comparison between JSMA, WJSMA and TJSMA}\label{tab3}
            \begin{tabular}{C{5cm}C{2cm}C{2cm}C{2cm}}
                \toprule
                \textbf{Attack} & \textbf{JSMA} & \textbf{WJSMA} & \textbf{TJSMA}\\
                \midrule
                \textbf{Time (second)} & 3964 & \textbf{3092} & \textbf{2797} \\
                \bottomrule
            \end{tabular}
            \label{speed}
        \end{table}

    We also notice that in this evaluation, the adversarial samples were crafted one by one. In practice, it is possible to generate samples by batch. In this case, the algorithm stops when all samples are finished. Most of the time, with a batch of large size, the three attacks approximately take the same time to converge. For example, on the same machine as previously and with a batch size equal to 1000, we were able to craft the same amount of samples in about 250s, for all the attacks.
    
    \textbf{Defense.} The objective now is to train neural networks so that the attacks fail as much as possible. One way of doing this is by adding adversarial samples crafted by JSMA, WJSMA and TJSMA to the training set. This method of training may imply a decrease in the model accuracy but adversarial examples will be more difficult to generate.
    
    We experiment this idea on the MNIST model in every possible configuration. To this end, 2,000 adversarial samples per class (20,000 more images in total), with distortion under 14.5\%, are added to the original MNIST training set, crafted by either JSMA, WJSMA or TJSMA. Then, three distinct models are trained on these augmented datasets. The models roughly achieve an accuracy of 99.9\% on the training set and 99.3\% on the test set, showing a slight loss compared to our previous MNIST model. Nevertheless, the obtained neural networks are more robust to the attacks as shown in Table \ref{defense}. Note that each experiment is made over the well-predicted samples of the test images. For each model and image, nine adversarial examples are generated by the three attacks.

    \begin{table}[ht]
        \centering
        \caption{Metrics (1) and (2) on JSMA, WJSMA and TJSMA augmented sets }
        \label{defense}
        \begin{tabular}{C{5.5cm}C{2cm}C{2cm}C{2cm}}
        \toprule
        \textbf{Metric} & \textbf{JSMA} & \textbf{WJSMA} & \textbf{TJSMA}\\
        \midrule
        \multicolumn{4}{c}{\textbf{Model trained over JSMA augmented set (9940 well predicted samples)}} \\
        \midrule
    	Success rate&77.94\%&\textbf{84.79\%}&\textbf{85.08\%}\\
        Mean $L_0$ distance on successful samples&54.48&\textbf{52.66}&\textbf{52.83}\\
        \midrule
        \multicolumn{4}{c}{\textbf{Model trained over WJSMA augmented set (9936 well predicted samples)}} \\
        \midrule
        Success rate&77.61\%&\textbf{90.05\%}&\textbf{92.01\%}\\
        Mean $L_0$ distance on successful samples&56.29&\textbf{52.72}&\textbf{52.18}\\
        \midrule
        \multicolumn{4}{c}{\textbf{Model trained over TJSMA augmented set (9991 well predicted samples)}} \\
        \midrule
        Success rate&76.42\%&\textbf{86.18\%}&\textbf{87.36\%}\\
       Mean $L_0$ distance on successful samples&54.26&\textbf{54.20}&\textbf{54.49}\\
        \bottomrule
        \end{tabular}
    \end{table}
    
    Overall, the attacks are less efficient on each of these models, compared to Table \ref{mnist}. The success rates drop by about 8 pp, whereas the number of iterations is increased by approximately 26\%. From the defender's point of view, networks trained against WJSMA and TJSMA give the best performance. The JSMA trained model provides the lowest success rates while the TJSMA trained network is more robust from the $L_0$ distance point of view. From the attacker's point of view, TJSMA remains the most efficient attack regardless of the augmented used dataset.

    \subsubsection{Comparison with CW $L_0$ attack.} Because of the complexity of this attack, comparison on a large number of images as before is very costly. For this reason, we only provide results on the first 100 well-predicted images of CIFAR-10, thus on 1,000 adversarial images given in Table \ref{speed1}. 
    
    	\begin{table}[]
    		\centering
    		\caption{Results for $L_0$ CW on CIFAR-10}\label{tab56}
    		\begin{tabular}{C{2cm}C{3cm}C{4cm}}
    			\toprule
    			\textbf{Success rate} & \textbf{Mean $L_0$ distance} & \textbf{Time}\\
    			\midrule
    			99.89\% & 24.97 &  On average more than one hour and a half to generate 9 adversarial samples run one by one on GPU \\
    			\bottomrule
    		\end{tabular}
    		\label{speed1}
    	\end{table}
    	
    We report better results of CW in terms of success rate and $L_0$ distance and a remarkable speed advantage of our attacks. Indeed, generating 9 adversarial samples (one by one) from a CIFAR-10 image by CW took on average near one hour and a half on GPU. The same task took 100 seconds for our attacks (without batching and $25$ seconds when batching). This makes our attacks at least $54$ times faster than CW $L_0$.
     
    \subsection{Experiments on non-targeted attacks}\label{weighted} 
\begin{figure}[H]
    	\centering
    	\includegraphics[width=0.8\textwidth]{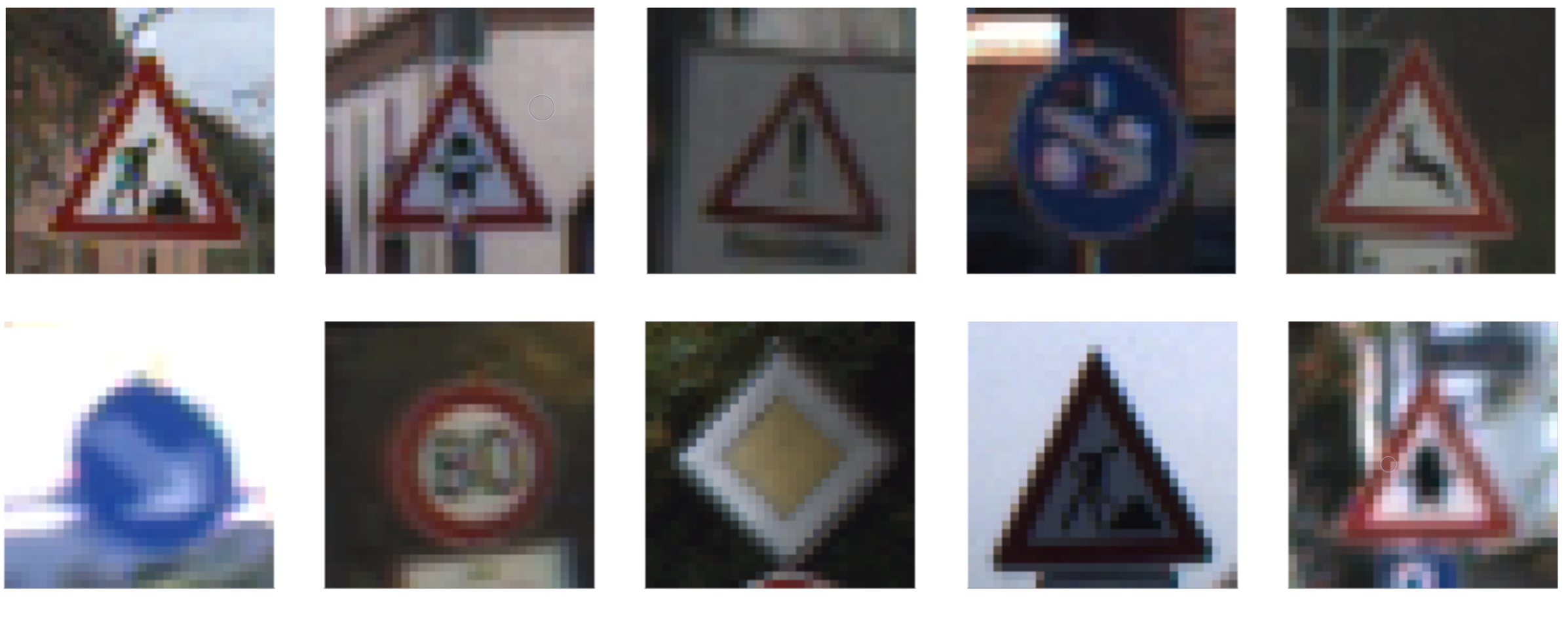}
    	\caption{Examples of images from GTSRB}
    	\label{gtrsbimages}
    \end{figure}

    In this part, we test the new NT attacks in the white/black-box modes and compare their performances with NT-JSMA and M-JSMA. In the white-box mode, we also compare with NT CW $L_0$. For experimentation, we chose the GTSRB dataset \cite{Stallkamp2012} widely used to challenge neural networks, especially in autonomous driving environments \cite{DBLP:journals/corr/PapernotMGJCS16}. We recall that GTSRB contains RGB $32\times 32\times 3$ traffic signs images, 86989 in the training set, and 12630 in the test set classified into 43 different possible categories. Figure \ref{gtrsbimages} displays some images from this dataset.
    
    \subsubsection{White-box experiments} 
    We consider a simplified NNC, described in Appendix \ref{appendix1}, whose architecture is inspired by Alexnet \cite{Krizhevsky_imagenetclassification}. After training this model with $\mathtt{cleverhans}$, it reaches 99.98\% accuracy on the training dataset and 95.83\% accuracy on the test set. 
    
    We implemented our attacks and those of \cite{JSMA2} again with TensorFlow using $\mathtt{cleverhans}$. During the first experiment, we run the attacks given in Table \ref{whitebox} on the first 1,000 test images after taking a maximum distortion $\gamma=3.7\%$ similar to CIFAR-10. The obtained results are shown in the same table. 
    
    	\begin{table}
    		\centering
    		\caption{Comparison of the performances between the NT attacks over the 1,000 first test images of GTSRB.} \label{whitebox}
    		\begin{tabular}{C{2cm}C{2cm}C{2cm}C{2cm}C{2cm}}
    			\toprule
    			\textbf{Attack} (U=JSMA) & \textbf{Success rate} & \textbf{$L_0$ average} & \textbf{Time (sec)}\\
    			\midrule
    			NT-U  & 91.35\% & 20.66 &  3604\\
    			NT-WU (ours) & \textbf{95.31\%} & \textbf{17.63} &  \textbf{760}\\
    			NT-TU (ours) & \textbf{96.87\%} & \textbf{15.86} &  \textbf{660}\\
    			\midrule
    			M-U & 98.44\% & 18.99 &7470\\
    			M-WU (ours) & \textbf{99.37\%} & \textbf{15.52} &  \textbf{6302}\\
    			\midrule
    			CW $L_0$  & 97.81\% & \textbf{13.56} & 260751\\
    			\bottomrule
    		\end{tabular}
    	\end{table}
    
    To analyse these results, it makes more sense to separately compare the NT-versions (faster), the M-versions (the most effective in success rate but slower) and discuss a global comparison with CW $L_0$. First, we notice a significant advantage of our NT versions over NT-JSMA according to the three metrics. In particular, NT-TJSMA is up to $5.4\times$ faster than NT-JSMA with nearly $4.8$ less modified pixels on average and more than 5 percentage points in success rate. We also notice that NT-TJSMA is the most effective attack among the three NT versions. Its notable benefit over NT-WJSMA is due to the fact that it converges in much less iterations than NT-WJSMA while both attacks need approximately the same time for an iteration. As for the M-versions, our attack M-WJSMA outperforms M-JSMA according to the three reported metrics. Both attacks are however slower than the NT versions. Finally, we notice that the gap between our attack M-WJSMA and CW $L_0$ is reduced as we obtain a better success rate, achieve less than two pixels in $L_0$ average while being more than 40 times faster. 
    
    \subsubsection{Black-box experiments} Here, we consider that black-box attack means the algorithm can use the targeted model as an oracle : when feeding it an image, the oracle returns a single class label. Moreover, we want to make only a limited amount of queries, which in a realistic setup would mean avoiding any suspicious activity, or at least trying not to depend too much on the oracle. To overcome this restriction, we train a substitute NNC using the Jacobian-based Dataset Augmentation (JBDA) \cite{DBLP:journals/corr/PapernotMGJCS16} method. We then perform white-box attacks on the substitute NNC. If it is good enough, the resulting adversarial images should \emph{transfer} to the oracle, i.e. effectively fool it even though they were designed to fool the substitute. Details about the substitute NNC architecture can be found in Appendix \ref{appendix1}. The JBDA method allows us to make queries to the oracle only during the training of the substitute network. Black-box attacks are thus closer to real-life setups, as one only needs to access for a short period of time the targeted model before being able to durably fool it. A dangerous application of such techniques is against autonomous driving cars and their sign recognition systems. NT black-box attacks are exactly the kind of threat that can be put in practice with relative ease and still disrupt considerably the car's behaviour. We will illustrate this insecurity through the GTSRB dataset.

    Attacks in this paragraph are NT-JSMA and M-JSMA \cite{JSMA2}, along with our contributions NT-WJSMA, NT-TJSMA and M-WJSMA. We experiment with different distortion rates. Contrary to white-box attacks, in the context of black-box attacks, the distortion rate is not an upper bound on the percentage of pixels that can be modified, but the exact percentage of pixels we want to perturb. This slight difference accounts for the imperfection of the substitute network : even if it mimics quite well the oracle, stopping as soon as the image switches according to the substitute will often not yield good results, so it is necessary to force the algorithm to push a little bit further. As a consequence, to evaluate the performance of the attacks, we will use two metrics : the success and transferability rates, for each distortion value. The first one measures the percentage of attacks that have been successful on the substitute NNC, while the second one measures the same percentage but for the oracle. The obtained results are given in Table \ref{blackbox}.
    
    	\begin{table}[H]
    		\centering
           \caption{Success and transferability rates (SR \& TR) in \% for the black-box NT attacks on the 1, 000 to 2, 000 test samples of GTSRB for a distortion $\gamma$ varying from 1\% to 5\%.}\label{blackbox}
    		\begin{tabular}{C{1.5em}C{3em}C{3em}C{3em}C{3em}C{3em}C{3em}C{3em}C{3em}C{3em}C{3em}C{3em}}
    			\toprule
    			\multirow{2}{*}{\textbf{$\gamma$}} &
    			\multirow{2}{*}{\textbf{Metric}} &
    			\multicolumn{5}{c}{\textbf{NT-Attack-Z (X=JSMA)}} &
    			\multicolumn{5}{c}{\textbf{NT-Attack-F  (X=JSMA)}} \\
    			&& X & WX & TX & M-X & M-WX & X & WX & TX & M-X & M-WX\\
    			\midrule
    			\multirow{2}{*}{1\%} & SR & 65.9 & 75.4 & 78.6 & 37 & 40.7 & 75.3 & 77.1 & \textbf{80.1} & 77 & 77.4 \\
    			& TR & 43 & 46.5 & 48.2 & 31.8 & 34.2 & 48.1 & 44.8 & \textbf{50.3} & 45.4 & 43.4 \\
    			\midrule
    			\multirow{2}{*}{2\%} & SR & 81.2 & 89.4 & 93.1 & 55.4 & 56.8 & 88.6 & 91.3 & 94.5 & 94.9 & \textbf{95.3}\\
    			& TR & 56.6 & 59.2 & 63.4 & 50.1 & 49.5 & 62.1 & 58 & \textbf{64.1} & 56.5 & 58.8\\
    			\midrule
    			\multirow{2}{*}{3\%} & SR & 88.5 & 95.5 & 97.7 & 68.9 & 67.4 & 94.4 & 95.8 & 98.6 & 99 & \textbf{99.2}\\
    			& TR & 65.1 & 67.2 & 70.2 & 58.8 & 59.2 & 69.3 & 64.1 & \textbf{70.7} & 64.5 & 65.3 \\
    			\midrule
    			\multirow{2}{*}{4\%} & SR & 92.9 & 97.1 & 99.2 & 77.7 & 74.7 & 97.4 & 98.3 & 99.5 & \textbf{100} & \textbf{100}\\
    			& TR & 69.2 & 71 & \textbf{75.3} & 68.5 & 65.1 & 73.9 & 67.5 & 74.8 & 69.9 & 69.5\\
    			\midrule
    			\multirow{2}{*}{5\%} & SR & 96 & 98.7 & 99.9 & 83.5 & 79.5 & 98.5 & 99 & \textbf{100} & \textbf{100} & \textbf{100}\\
    			& TR & 71.9 & 73.6 & \textbf{77.9} & 72.7 & 68.4 & 76.3 & 71.1 & 77.2 & 72.3 & 72.1\\
    			\bottomrule
    		\end{tabular}

    	\end{table}

    We can see in this table that in terms of success rate, the results are compatible with the white-box attack results, meaning that M-WJSMA mostly outperforms NT-TJSMA, which is better than NT-WJSMA which beats NT-JSMA, at least for the F variants. This is somewhat expected, because the attacks as performed on the substitute are merely white-box. However, one can notice that the Z variants of Maximal attacks do not perform well on this substitute network. More interestingly, concerning the transferability of the attacks, one can notice that for the Z attacks, the same hierarchy NT-TJSMA $>$ NT-WJSMA $>$ NT-JSMA is respected, while for the F attacks, NT-WJSMA is overall inferior to all the other attacks, and NT-TJSMA outperforms NT-JSMA, M-JSMA and M-WJSMA. 
    
    Overall, NT-TJSMA is the best attack for black-box non-targeted purposes, but the best variant (F or Z) depends on the distortion rate : NT-TJSMA-Z only beats its counterpart NT-TJSMA-F for $\gamma=4$ or $5$ \%. Finally, if one variant were to be chosen, it would be NT-TJSMA-F due to its speed and overall best transferability. 
    
    \section{Comparisons with non-$L_0$ attacks and conclusion}\label{conclusion}
    
    \subsection{Comparison with non-$L_0$ attacks}
    Previously, we only compared with $L_0$ attacks as it makes more sense to consider  methods that optimise the same metric. In this section, we compare our NT-TJSMA with a well-known non-targeted $L_{\infty}$ attack which is the Fast Gradient Sign Method (FGSM) \cite{FGM}. We recall that FGSM  attempts to minimise the $L_{\infty}$ norm. It is a very fast method; significantly faster than NT-TJSMA. To this end, we run both attacks by dropping the assumption on the number of modified input features for our attack and by experimenting with different values of the $L_{\infty}$ threshold $\varepsilon$ for FGSM where the results of the best threshold are kept. Then, we computed the mean $L_1$ and $L_2$ errors for each attack as alternative comparison metrics. Table \ref{f_table} shows the obtained results.
    
    \begin{table}[H]
        \centering
        \caption{Comparison between NT-TJSMA and FGSM}
        \label{fgsm_comp}
        \begin{tabular}{C{2.5cm}C{3cm}C{3cm}C{3cm}}
        \toprule
        \textbf{Metric} & \textbf{MNIST} & \textbf{CIFAR10} & \textbf{GTSRB}\\
        \midrule
        \multicolumn{4}{c}{\textbf{Performances of NT-TJSMA}} \\
        \midrule
    	Success rate & 100\% & 100\% & 100\% \\
        Mean $L_1$ distance & 13.58 & 13.33 & 15.01 \\
        Mean $L_2$ distance & 3.49 & 2.93 & 2.88 \\
        \midrule
        \multicolumn{4}{c}{\textbf{Performances of FGSM}} \\
        \midrule
        Success rate & 93.2\% ($\varepsilon=0.75$) & 88.2\% ($\varepsilon=0.05$) & 99.5\% ($\varepsilon=0.9$) \\
        Mean $L_1$ distance & 227.21 & 150.85 & 1522.31 \\
        Mean $L_2$ distance & 12.80 & 2.74 & 32.54 \\
        \bottomrule
        \end{tabular}
    \end{table}\label{f_table}
    
    \noindent As it can be seen, NT-TJSMA is always successful for each model, while FGSM is far from reaching $100\%$ SR. Moreover, NT-TJSMA obtains better $L_1$ and $L_2$ scores. Thus our attack outperforms FGSM for the SR and the $L_0, L_1$ and $L_2$ metrics, while FGSM has only the $L_{\infty}$ and speed advantages. Note that for FGSM we considered the best results for different thresholds, while our attack is run one time.
    \subsection{Conclusion}
    In this section, we summarise our main findings and also discuss the limitation of our work.

    We have introduced WJSMA and TJSMA, new probabilistic adversarial variants of JSMA for targeted and non-targeted misclassification of deep neural network classifiers.
    
    Experiments in the targeted case have demonstrated, after analysing a large amount of images (more than 790,000 images), that our targeted attacks are more efficient and also faster than JSMA. It is important to recall the quite natural derivation of these attacks from a simple and classical log softmax reasoning which has not been noticed before. 
    Our attacks do not beat CW $L_0$ but have an important speed advantage highlighted in the paper (more than 50 times faster on CIFAR-10). Therefore, for targeted $L_0$ misclassification, they offer substantial tools to test neural networks in real-time. This fact is supported by our fast implementation provided with the paper.
    
    As a second contribution, we have introduced NT and M variants of WJSMA/TJSMA and have shown that they outperform the previous NT and M versions of JSMA. Through experiments on GTSRB, we noticed that the gap between our attacks and CW in $L_0$ average is reduced. Moreover, we obtained better success rates, while remaining at least 40 times faster than CW $L_0$.
    
    In the NT part of the paper, we did not compare our attacks with the one pixel attack \cite{DBLP:journals/tec/SuVS19}. Indeed, this approach has a high computational cost and we only claim an advantage in speed which is quite evident for us (see also the time evaluation in \cite{DBLP:conf/cvpr/ModasMF19}). Also, we did not provide a comparison with SparseFool \cite{DBLP:conf/cvpr/ModasMF19} an effective NT $L_0$ attack because of the need to reimplement this attack with TensorFlow. On CIFAR-10, \cite{DBLP:conf/cvpr/ModasMF19} found that crafting an example by SparseFool takes on average 0.34 and 0.69 second on two different neural networks. Our speed performances are very competitive with these values. Indeed, 
    on GTSRB which has many more classes than CIFAR-10, our NT-TJSMA was able to craft an example in near 0.68 seconds on average (counting only successful images). Thus, regarding SparseFool, we first claim competitive results in speed. Second, our results obtained on LeNet-5 (more than 99.5\% on a model similar to \cite{DBLP:conf/cvpr/ModasMF19}, see Section \ref{confusion}) are very close to \cite{DBLP:conf/cvpr/ModasMF19} although we only run the attacks up to a limited maximum number of iterations contrary to \cite{DBLP:conf/cvpr/ModasMF19}.
    
    Overall, our results suggest that for adversarial purposes, TJSMA, M-WJSMA and NT-TJSMA should be preferred over the original variants of JSMA, respectively in the case of white-box targeted attacks, white-box non-targeted attacks, and black-box non-targeted attacks. We recall that despite the fact that TJSMA is a more elaborate version of WJSMA, it is hardly compatible with the ``Maximal'' approach, which in turn proves to be very efficient for non-targeted purposes. For this reason, as we have demonstrated, M-WJSMA is indeed the right choice for this type of attacks. On the other hand, because the ``Maximal'' approach has not proved to be very efficient on black-box non-targeted attacks, it is the non-targeted version of TJSMA (NT-TJSMA) that is the best in this case.
    
    Finally, we should mention that despite improving JSMA in different ways, like JSMA, our approach is still not scalable to large datasets. This is because of the high computational cost of saliency maps when the dimension of inputs becomes large. Our approach is therefore intended for ``small" datasets such as those considered in the paper. Nevertheless, this kind of datasets is very common in real-life applications. See also the recent paper  \cite{2020arXiv201112423H}.

    \vspace{6pt} 
    \textbf{Supplementary Material}
    All our codes are publicly available through the link \url{https://github.com/probabilistic-jsmas/probabilistic-jsmas}.\label{supp}
    
    
    \textbf{Author contributions:}
    Conceptualization: T.C., A.L., M.F. and H.H.; Software: T.C., A.L., M.F. and H.H.; Data curation: T.C., A.L. and M.F.; Methodology: A.L. and H.H.; Supervision: H.H., Writing - original draft : T.C., A.L., M.F. and H.H.; Writing - review and editing : T.C., A.L., M.F. and H.H. All authors have read and agreed to the published version of the manuscript.
    
    
    \textbf{Funding:}
    This research received no external funding.
    
    \textbf{Acknowledgments:}
    This work was done in the context of an internship by T. Combey, A. Loison and M. Faucher supervised by H. Hajri. We thank Gabriel Zeller for his assistance. We are grateful to Wassila Ouerdane and Jean-Philippe Poli at CentraleSup\'elec for their support. We thank the mesocentre de calcul Fusion, Metz computing center of CentraleSup\'elec and St\'ephane Vialle for providing us effective computing resources. H. Hajri is grateful to Sylvain Lamprier for useful discussions, the scientific direction and the EPI project (\'Evaluation des Performances de syst\`emes de d\'ecision \`a base d'Intelligence Artificielle) at IRT SystemX for their support.
    
    \textbf{Conflicts of interest:}
    The authors declare no conflict of interest.
    
    \textbf{Abbreviations:}
    The following abbreviations are used in this manuscript:\\
    
    \noindent 
    \centering
    \begin{tabular}{@{}ll}
    NNC & Neural Network Classifier \\
    JSMA & Jacobian-based Saliency Map Attack\\
    MJSMA & Maximal Jacobian-based Saliency Map Attack\\
    WJSMA & Weighted Jacobian-based Saliency Map Attack\\
    TJSMA & Taylor Jacobian-based Saliency Map Attack\\
    NT & Non-Targeted\\
    M & Maximal\\
    CW & Carlini-Wagner
    \end{tabular}
    
    \appendix
    \section*{APPENDIX}
    \section{Architectures of the deep NNCs.}\label{appendix1}
    
    \begin{table}[H]
    \centering
    \caption{Architecture of the used NNC on MNIST (LeNet-5 inspired)}
    \label{mnistarchi}
    \resizebox{250pt}{!}{%
    \begin{tabular}{|c|c|}
     \hline
     \textbf{Layer} & \textbf{Parameters} \\
     \hline
     Input Layer & size: ($28\times 28$)\\
     \hline
     Conv2D  & kernel size: ($5\times 5$), $20$ kernels, no stride \\
     \hline
     ReLu   & \\
     \hline
     MaxPooling2D 
          & kernel size: ($2\times 2$), stride: ($2\times 2$)\\
     \hline
     Conv2D  & kernel size: ($5\times 5$), $50$ kernels, no stride \\
     \hline
     ReLu& \\
     \hline
     MaxPooling2D 
          & kernel size: ($2\times 2$), stride: ($2\times 2$)\\
     \hline
     Flatten  &\\
     \hline
     Dense   & size: $500$ \\
     \hline
     ReLu   &\\
     \hline
     Dense   & size: number of classes (10 for MNIST)\\
     \hline
     Softmax  &\\
     \hline
    \end{tabular}}
    \end{table}
    
    \begin{table}[H]
    \centering
    \caption{Architecture of the used NNC on CIFAR-10}
    \label{cifar10nn}
    \resizebox{250pt}{!}{%
    \begin{tabular}{|c|c|}
     \hline
     \textbf{Layer} & \textbf{Parameters} \\
     \hline
     Input Layer & size: ($32\times 32 \time 3$)\\
     \hline
     Conv2D  & kernel size: ($3\times 3$), $64$ kernels, \newline no stride \\
     \hline
     ReLu   & \\
     \hline
     Conv2D  & kernel size: ($3\times 3$), $128$ kernels, no stride \\
     \hline
     ReLu   & \\
     \hline
     MaxPooling2D 
          & kernel size: ($2\times 2$), stride: ($2\times 2$)\\
     \hline
     Conv2D  & kernel size: ($3\times 3$), $128$ kernels, no stride \\
     \hline
     ReLu& \\
     \hline
     Conv2D  & kernel size: ($3\times 3$), $256$ kernels, no stride \\
     \hline
     ReLu& \\
     \hline
     MaxPooling2D 
          & kernel size: ($2\times 2$), stride: ($2\times 2$)\\
     \hline
     Conv2D  & kernel size: ($3\times 3$), $256$ kernels, no stride \\
     \hline
     ReLu& \\
     \hline
     Conv2D  & kernel size: ($3\times 3$), $512$ kernels, no stride \\
     \hline
     ReLu& \\
     \hline
     MaxPooling2D 
          & kernel size: ($2\times 2$), stride: ($2\times 2$)\\
     \hline
     Conv2D  & kernel size: ($3\times 3$), $10$ kernels, no stride \\
     \hline
     GlobalAveragePooling 
          & kernel size: ($2\times 2$), stride: ($2\times 2$)\\
     \hline
     Softmax  &\\
     \hline
    \end{tabular}%
    }
    \end{table}
    
    \begin{table}[H]
    	\centering
    	\caption{Architecture of the used NNC on GTSRB (AlexNet inspired)}
    	\label{cifar10nn}
    	\begin{tabular}{|c|c|}
    		\hline
    		\textbf{Layer} & \textbf{Parameters} \\
    		\hline
    		Input Layer & size: ($32\times 32 \time 3$)\\
    		\hline
    		Conv2D  & kernel size: ($5\times 5$), $64$ kernels, \newline no stride \\
    		\hline
    		ReLu   & \\
    		\hline
    		MaxPooling2D & kernel size: ($3\times 3$), stride: ($2\times 2$)\\
    		\hline
    		Conv2D  & kernel size: ($5\times 5$), $64$ kernels, \newline no stride \\
    		\hline
    		ReLu   & \\
    		\hline
    		MaxPooling2D & kernel size: ($3\times 3$), stride: ($2\times 2$)\\
    		\hline
    		Flatten   & \\
    		\hline
    		Dense & size: $384$\\
    		\hline
    		ReLu & \\
    		\hline
    		Dense & size: $192$ \\
    		\hline
    		ReLu & \\
    		\hline
    		Dense & size: $43$ \\
    		\hline
    		Softmax  &\\
    		\hline
    	\end{tabular}
    \end{table}
    
    \begin{table}[H]
    	\centering
    	\caption{Architecture of the used substitute NNC on GTSRB}
    	\label{subnn}
    	\begin{tabular}{|c|c|}
    		\hline
    		\textbf{Layer} & \textbf{Parameters} \\
    		\hline
    		Input Layer & size: ($32\times 32 \times 3$)\\
    		\hline
    		Conv2D  & kernel size: ($3\times 3$), $16$ kernels, \newline no stride \\
    		\hline
    		ReLu   & \\
    		\hline
    		MaxPooling2D & kernel size: ($2\times 2$), stride: ($2\times 2$)\\
    		\hline
    		Conv2D  & kernel size: ($3\times 3$), $32$ kernels, \newline no stride \\
    		\hline
    		ReLu   & \\
    		\hline
    		MaxPooling2D & kernel size: ($2\times 2$), stride: ($2\times 2$)\\
    		\hline
    		Conv2D  & kernel size: ($3\times 3$), $64$ kernels, \newline no stride \\
    		\hline
    		ReLu   & \\
    		\hline
    		Flatten   & \\
    		\hline
    		Dense & size: $43$\\
    		\hline
    		Softmax  &\\
    		\hline
    	\end{tabular}
    \end{table}
    
    \section{Pairwise dominance.}\label{appendix2}
    \begin{figure}[H]
    	\centering
    	\begin{subfigure}[b]{0.46\textwidth}
    			    \caption{WJSMA vs JSMA}
    			    \label{sub11}
    			    \vspace{0.2cm}
    	    \includegraphics[width=\textwidth]{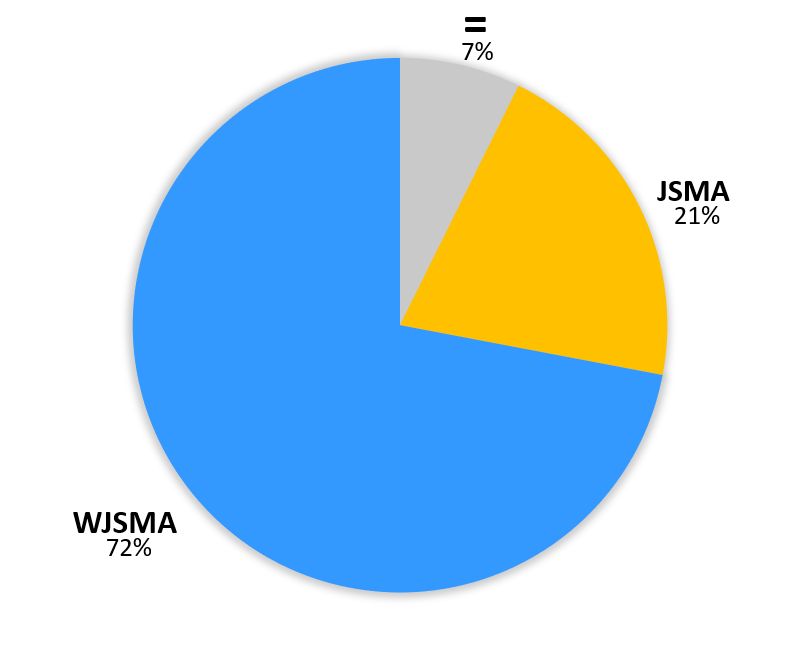}
    
    	    \label{jsma_wjsma_mnist}
    	\end{subfigure}
    	\begin{subfigure}[b]{0.46\textwidth}
    			    \caption{TJSMA vs JSMA}
    			    \label{sub21}
    			    			    \vspace{0.2cm}
    	    \includegraphics[width=\textwidth]{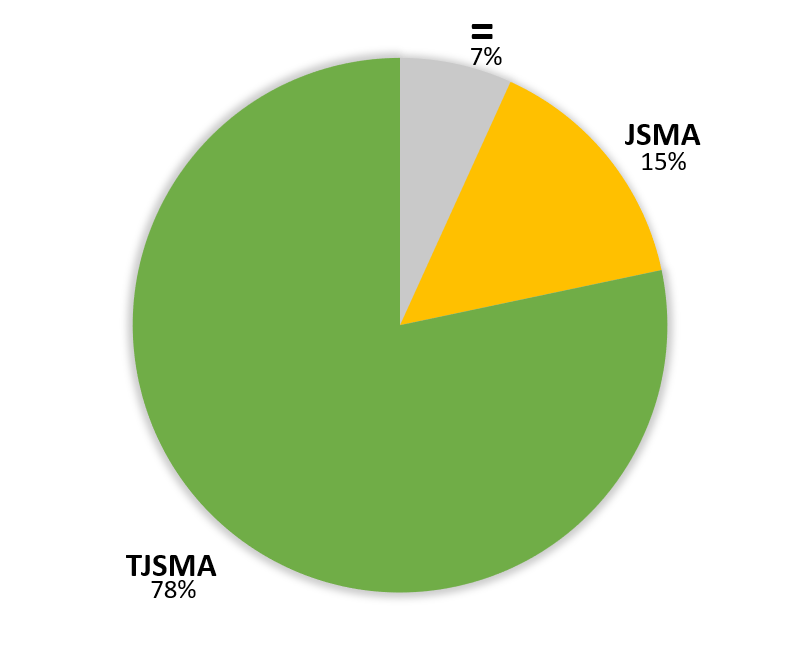}
    
    	    \label{jsma_tjsma_mnist}
    	\end{subfigure}
    \vspace{0.3cm}
    	\begin{subfigure}[b]{0.46\textwidth}
    	    \includegraphics[width=\textwidth]{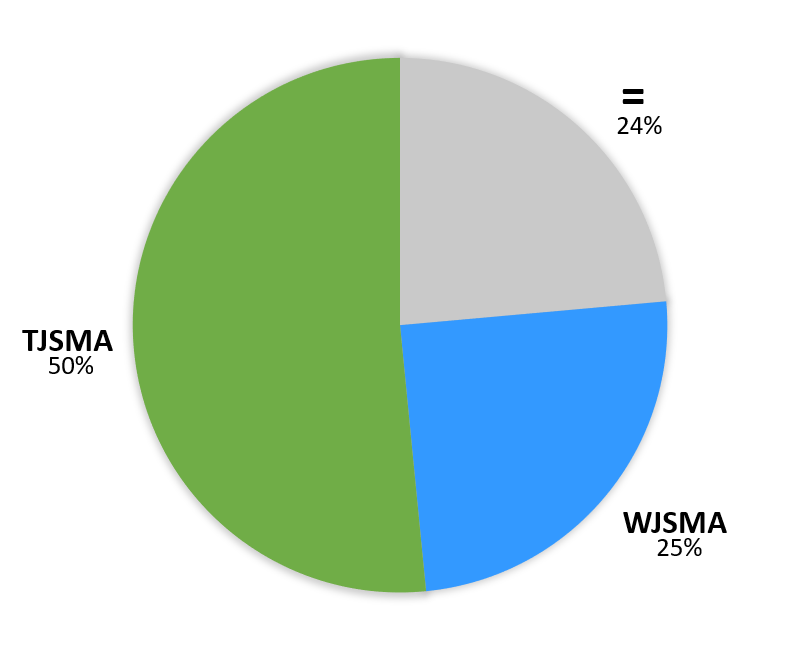}
    	    \caption{TJSMA vs WJSMA}
    	    \label{sub31}
    	    \label{wjsma_tjsma_mnist}
    	\end{subfigure}
    	\caption{Pairwise dominance on MNIST comparing WJSMA with JSMA (\ref{sub11}), TJSMA with JSMA (\ref{sub21}) and TJSMA with WJSMA (\ref{sub31}). On these charts, "=" corresponds to samples with the same number of iterations by the attacks including when both attacks fail.  }
    	\label{dominances_mnist}
    \end{figure}
    
    \begin{figure}[H]
    	\centering
    	\begin{subfigure}[b]{0.46\textwidth}
    			    \caption{WJSMA vs JSMA}
    			    \vspace{0.2cm}
    	    \includegraphics[width=\textwidth]{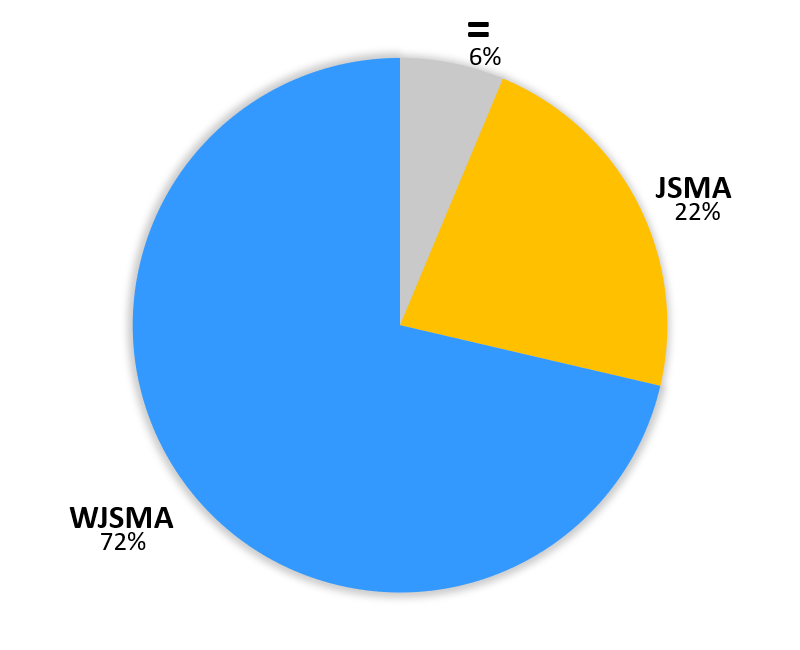}
    
    	    \label{jsma_wjsma_cifar}
    	\end{subfigure}
    	\begin{subfigure}[b]{0.46\textwidth}
    			    \caption{TJSMA vs JSMA}
    			    \vspace{0.2cm}
    	    \includegraphics[width=\textwidth]{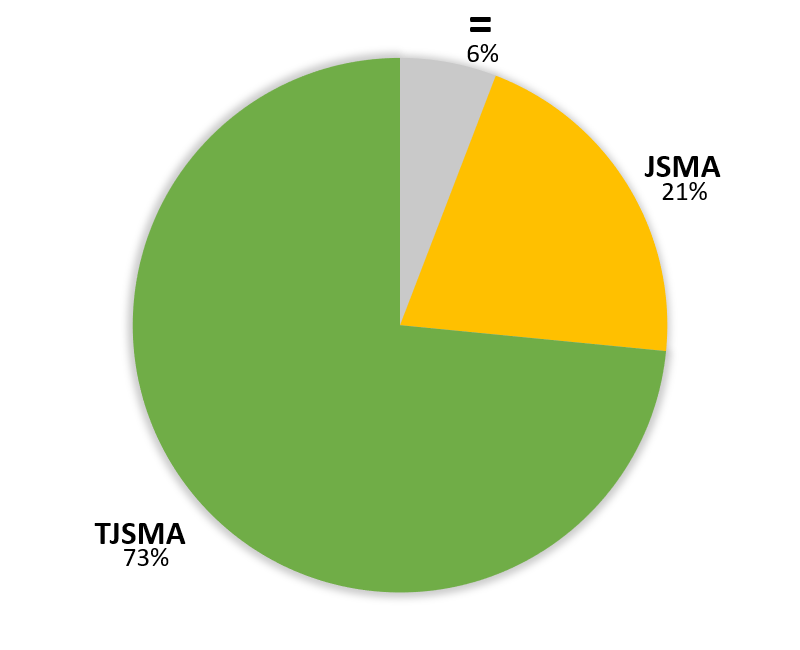}
    
    	    \label{jsma_tjsma_cifar}
    	\end{subfigure}
    	\begin{subfigure}[b]{0.46\textwidth}
    	    \includegraphics[width=\textwidth]{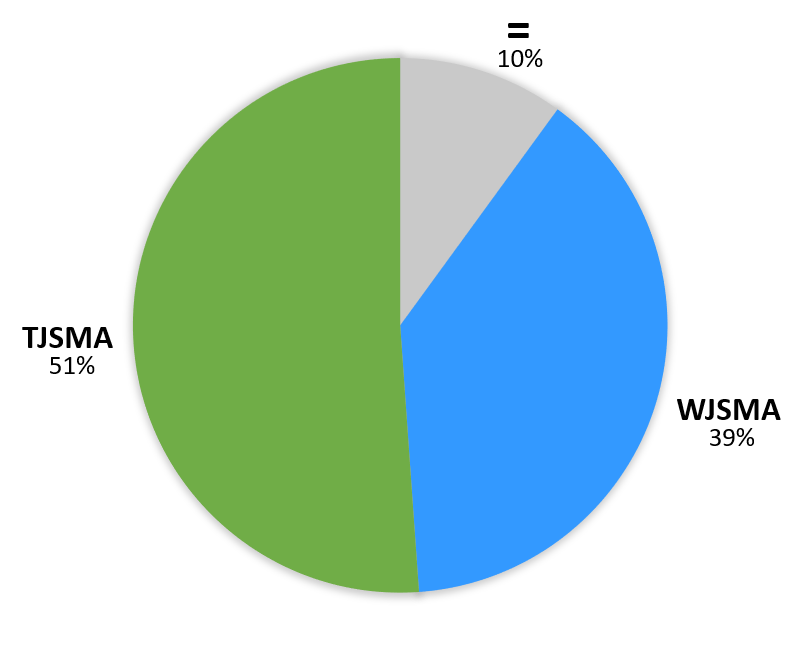}
    	    \caption{TJSMA vs WJSMA}
    	    \label{wjsma_tjsma_cifar}
    	\end{subfigure}
    	\caption{Pairwise dominance on CIFAR-10  comparing WJSMA with JSMA (\ref{jsma_wjsma_cifar}), TJSMA with JSMA (\ref{jsma_tjsma_cifar}) and TJSMA with WJSMA (\ref{wjsma_tjsma_cifar}). "=" has the same significance as before.}
    	\label{dominances_cifar}
    \end{figure}

    Further analysis of the results obtained on MNIST reveals that, even for examples where JSMA is better than WJSMA or TJSMA, on average, less than 10 more components are changed by WJSMA or TJSMA, whereas JSMA changes more than 17 more components on average when it is dominated by WJSMA or TJSMA. A similar gap can be noticed on CIFAR-10.
    
    
    \bibliographystyle{plain}
    \bibliography{main}
    \end{document}